\documentclass[letterpaper]{article} 
\usepackage{aaai25}  
\usepackage{times}  
\usepackage{helvet}  
\usepackage{courier}  
\usepackage[hyphens]{url}  
\usepackage{graphicx} 
\usepackage{natbib}  
\usepackage{caption} 
\usepackage{algorithm}
\usepackage{algpseudocode}
\usepackage{newfloat}
\usepackage{listings}
\DeclareCaptionStyle{ruled}{labelfont=normalfont,labelsep=colon,strut=off} 
\floatstyle{ruled}
\newfloat{listing}{tb}{lst}{}
\floatname{listing}{Listing}
\nocopyright 

\title{On the Robustness of Distributed Machine Learning against\\ Transfer Attacks}
\author{
     Sébastien Andreina\textsuperscript{\rm 1}, Pascal Zimmer\textsuperscript{\rm 2}, Ghassan Karame\textsuperscript{\rm 2}
}
\affiliations{
    \textsuperscript{\rm 1}NEC Labs Europe, Germany\\
    \textsuperscript{\rm 2}Ruhr University Bochum, Germany\\
    sebastien.andreina@neclab.eu, \{pascal.zimmer, ghassan.karame\}@rub.de  
}

\newif\iflongversion  
\longversiontrue      

\usepackage[T1]{fontenc}

\usepackage{lmodern}
\usepackage{microtype}

\usepackage{comment}
\usepackage{color}

\usepackage{url}

\usepackage[cmex10]{amsmath}
\usepackage{subcaption}

\usepackage{amssymb}
\usepackage{amsmath}

\usepackage{xspace}

\usepackage{booktabs}
\usepackage{multirow}
\usepackage{adjustbox}

\usepackage[inline]{enumitem}
\usepackage{siunitx}

\usepackage{cleveref}
\usepackage{nicefrac}

\definecolor{burgundy}{rgb}{0.5, 0.0, 0.13}
\definecolor{cadmiumgreen}{rgb}{0.0, 0.42, 0.24}
\definecolor{burntorange}{rgb}{0.8, 0.33, 0.0}
\definecolor{darkgray}{rgb}{0.66, 0.66, 0.66}
\definecolor{turquoise}{rgb}{0.19, 0.84, 0.78}
\definecolor{lightgray}{rgb}{0.77, 0.77, 0.77}
\definecolor{chamoisee}{rgb}{0.63, 0.47, 0.35}
\definecolor{carolinablue}{rgb}{0.6, 0.73, 0.89}

\DeclareMathOperator*{\argmax}{arg\,max}

\newcommand{\RR}{\mathbb{R}}
\newcommand{\size}[1]{\lvert #1 \rvert} 

\newcommand{\setX}{\mathcal{X}}
\newcommand{\setY}{\mathcal{Y}}

\newcommand{\adv}{\mathcal{A}} 
\newcommand{\dist}[1]{\lVert #1 \rVert} 

\newcommand{\advsuccrate}{\ensuremath{\mathsf{ASR}}\xspace}
\newcommand{\robustacc}{\ensuremath{\mathsf{RA}}\xspace}
\newcommand{\mainacc}{\ensuremath{\mathsf{CA}}\xspace}

\usepackage{array}
\newcolumntype{H}{>{\setbox0=\hbox\bgroup}c<{\egroup}@{}}

\newtheorem{proposition}{Proposition}

\definecolor{satinsheengold}{rgb}{0.8, 0.63, 0.21}

\usepackage{graphicx}

\newcommand{\Nodes}{\ensuremath{N}\xspace}
\newcommand{\Arch}{\ensuremath{A}\xspace}

\newcommand{\Optim}{\ensuremath{O}\xspace}
\newcommand{\Sched}{\ensuremath{S}\xspace}
\newcommand{\Momen}{\ensuremath{\nu}\xspace}

\newcommand{\Lr}{\ensuremath{\mu}\xspace}

\newcommand{\Wd}{\ensuremath{\lambda}\xspace}

\newcommand{\majorParam}{\ensuremath{\mathcal{P}}\xspace}
\newcommand{\hyperParam}{\ensuremath{\mathcal{H}}\xspace}

\begin{document}

\maketitle

\begin{abstract}
Although distributed machine learning (distributed ML) is gaining considerable attention in the community, prior works have independently looked at instances of distributed ML in either the training or the inference phase. No prior work has examined the combined robustness stemming from distributing both the learning and the inference process.

In this work, we explore, for the first time, the robustness of distributed ML models that are fully heterogeneous in training data, architecture, scheduler, optimizer, and other model parameters. Supported by theory and extensive experimental validation using CIFAR10 and FashionMNIST, we show that such properly distributed ML instantiations achieve across-the-board improvements in accuracy-robustness tradeoffs against state-of-the-art transfer-based attacks that could otherwise not be realized by current ensemble or federated learning instantiations. For instance, our experiments on CIFAR10 show that for  the Common Weakness attack, one of the most powerful state-of-the-art transfer-based attacks, our method improves robust accuracy by up to 40\%, with a minimal impact on clean task accuracy.\footnote{The code is available at \url{https://github.com/RUB-InfSec/distributed_learning_robustness}.}

\end{abstract}

\section{Introduction}
Nowadays, many applications increasingly rely on various forms of distributed learning. For instance, Google spell-checking utilizes horizontal federated learning~\cite{zhang2023privatefederatedlearninggboard}, and Apple, among others, may also be adopting similar approaches~\cite{paulik2021federatedevaluationtuningondevice}. Autonomous vehicles are soon expected to follow suit{~\cite{10316635}}. In the financial sector, vertical federated learning is currently employed~\cite{10415268}, and many researchers have already begun exploring decentralized learning without a central authority~\cite{decentralizedLearning,decentralizePy}.

Although federated and decentralized learning is gaining considerable interest and attention from practitioners and researchers, the concept of distributed learning was first attempted with ensemble learning. Here, weak learners employing different model architectures (typically co-located on the same machine) jointly train using the same dataset; inference requires some sort of majority vote among those weak learners. 
Unlike ensemble learning, federated learning typically requires weak learners to jointly train a global model (optionally with the help of a central server). The main motivation of these instantiations is to obviate the need for weak learners to share their local training data, hence increasing learners' privacy while allowing for diverse data to be effectively used for training. 
As such, current federated learning approaches primarily aim to diversify data sources for training but do not directly alter the inference process. In contrast, ensemble learning focuses mainly on inference by enhancing model diversity. 
The literature features a number of contributions that analyze the privacy provisions of FL and its resistance to backdoors on the one hand and the robustness offered by ensemble learning on the other hand. \emph{No prior work has examined the combined robustness and security provisions stemming from distributing both the learning and the inference process.} This is particularly relevant since centralized ML models, such as deep neural networks, have been shown to lack robustness against adversarial examples~\cite{DBLP:journals/corr/SzegedyZSBEGF13,DBLP:conf/pkdd/BiggioCMNSLGR13}.

In this paper, we address this gap and explore whether pure \emph{distributed learning}, a variant instantiation combining the benefits of both federated/decentralized learning and traditional ensemble learning, can provide increased robustness against transfer attacks.
We specifically focus on transfer attacks because they can be executed by resource-constrained adversaries that do not need to know the model parameters to create a surrogate model and conduct transferable white-box attacks.
More specifically, we consider a setting where weak learners can choose their dataset for training. However, unlike FL, learners can also independently choose their training (such as optimizer and scheduler) and model (i.e., architecture) parameters. We argue that such a hybrid setting allows us to analyze, for the first time, the robustness of distributed ML models that are heterogeneous in \emph{both} their training data \emph{and} training and model parameters. 
We contrast this to previous work that has independently looked at instances of ML in each isolated phase. 
For example, \cite{kingpaper} demonstrated that ensembles of models are more robust to certain attacks compared to single models. Similarly, \cite{DBLP:conf/uss/DemontisMPJBONR19} highlighted the significance of model architecture on the transferability of adversarial examples, and \cite{DBLP:conf/aaai/00810CL0D023} explored the robustness of FL frameworks, particularly focusing on the challenges of employing FL-based adversarial training in non-IID data settings. In contrast, we aim to address the following research questions in this work:
\begin{description}
	\item[RQ1] To what extent is distributed ML more robust than existing traditional ensemble learning against transfer-based attacks? 
	\item[RQ2] Are there specific model parameters of particular relevance to increase robustness in a distributed setting?
	\item[RQ3] How does the training data distribution between different distributed models impact the overall robustness?
	\item[RQ4] How do distributed aggregation schemes affect the robustness?
\end{description}
We conduct an extensive robustness evaluation of our approach with state-of-the-art transfer-based attacks and find across-the-board improvements in robustness against all considered attacks. For instance, our experiments on CIFAR10 show that for CW~\cite{chen2024rethinking}, one of the most powerful state-of-the-art transfer attacks, our method improves robust accuracy by up to 33.6\% and 41.2\%, with a minimal impact on clean task accuracy of at most between 1.1\% and 13.5\%.

\section{Background \& Related Work}
\subsection{ML Paradigms}
\noindent \textbf{``Centralized'' Learning} 
is the de-facto standard learning paradigm, where a single entity possessing a dataset trains a single model on it. While multiple servers can be used to improve the learning process speed, they typically all run the same code based on the same parameters and synchronize the results of the training at each epoch.

\vspace{0.25 em}\noindent \textbf{Ensemble Learning }
is a method that uses multiple learning algorithms and architectures to obtain better predictive performances compared to its centralized counterparts~\cite{ensemble-robustness}. 

\vspace{0.25 em}\noindent \textbf{Federated and Decentralized Learning} 
are settings where the complete training data is unavailable to a single entity due to privacy concerns. 
Both paradigms distribute the training process across multiple nodes, utilizing locally available data. 
Federated learning relies on a trusted centralized party to coordinate the training process and aggregate the models computed locally by each node. In contrast, decentralized learning eliminates the central trusted party by leveraging peer-to-peer communication to share models among the nodes.

\subsection{Transfer-based attacks}

While some early techniques required complete access to the target classifier to generate an adversarial example, recent research showed the feasibility of \emph{transferring} adversarial examples in a so-called ``blackbox'' setting~\cite{Dong_2018_CVPR, odi, Wang_2021_CVPR,linbp, Wang_2021_ICCV,kingpaper}. 
Here, the adversary does not have access to the model parameter but still has complete knowledge of the training parameters, such as architecture, dataset, hyperparameters, etc. 
The adversary can then train a substitute model and use it to generate adversarial examples that are likely to be misclassified by the target model. In this case, the adversary is not required to have access to the trained model.

Given a target classifier $C_t$ and a local ``surrogate'' classifier $C_s$, an attacker can simply run the standard white-box attack on surrogate classifier $C_s$ to generate an adversarial example that is likely to transfer to the target classifier $C_t$. 
Most previous work assumes the surrogate classifier $C_s$ has been trained using the same training data as the target classifier $C_t$~\cite{DBLP:journals/corr/GoodfellowSS14,odi,naa,linbp,paagaa}.
The transferability of the attack is then evaluated between different source and target classifier architectures trained with the same training set.

While adversarial examples generally transfer, the success rate of transfer-based attacks using plain standard white-box attacks is limited. To solve this shortcoming, there have been numerous works focused on improving the transferability of adversarial examples; \cite{Dong_2018_CVPR} improved transferability of FGSM~\cite{DBLP:journals/corr/GoodfellowSS14} by adding momentum to the gradient; \cite{xie_improving_2019} uses input diversity using random transform to improve the resilience of the adversarial transform.
More recent work, such as~\cite{taig,ttp,odi,naa,linbp,paagaa} use more involved techniques and achieve transferability very close to 100\% on undefended models.

\subsection{Related Work}
Previous work~\cite{DBLP:conf/uss/DemontisMPJBONR19,DBLP:conf/nips/InkawhichLWICC20} explored the transferability of adversarial examples but did not analyze the impact of training parameters on the robustness of decentralized deployments.
Additionally, the generalization of \cite{DBLP:conf/uss/DemontisMPJBONR19} is restricted to binary classifiers, does not include attacks optimized for transferability (focusing only on white box attacks), and does not investigate the impact of parameters beyond the architecture.
On the other hand, while \cite{beware} analyzed the impact of the training data over transferability, they did not consider other model parameters. 

Additionally, previous work outlined the improvement in robustness provided by ensemble learning~\cite{kurakin2018adversarial} over single models. \cite{DBLP:conf/icml/PangXDCZ19} improved the robustness of ensemble models by devising a strategy to increase the diversity of the trained models in ensemble training, reducing the transferability and yielding enhanced robustness. 

We conclude that previous work has looked at the isolated impact of a given (training) parameter on transferability. No prior work has
examined the combined robustness stemming
from distributing both the learning and the inference
process and the impact of the various hyperparameters on transferability.

\section{Methodology}

\subsection{Preliminaries and Notations}
We denote the sample and label spaces with $\mathcal{X}$ and $\mathcal{Y}$, respectively, and the training data with $\mathcal{D}=(x_i,y_i)^N_{i=1}$, where $N$ is the training set size.
A DNN-based classifier $f_{\theta}: \mathcal{X}\to[0,1]^{|\mathcal{Y}|}$ is a function (parameterized by $\theta$) that, given an input $x$ outputs the probability that the input is classified as each of the $n=|\mathcal{Y}|$ 
 classes. The highest prediction probability in this vector, i.e., $\max_{i\in[n]}f_{\theta,i}(x)$, is also called the \emph{confidence} of the model in the classification of the sample. The prediction of the classifier can be derived as $y=C(x):=\argmax_{i \in [n]}({f_{\theta,i}}(x))$.

We define an adversarial example $x'$ as a genuine image $x$ to which carefully crafted adversarial noise is added, i.e., $x' = x + \zeta$ for a small perturbation~$\zeta$ such that~$x'$ and~$x$ are perceptually indistinguishable to the human eye and yet are classified differently.

Given a genuine input~$x_0\in \RR^d$ predicted as~${C(x_0) = s}$ (source class), $x'$ is an \emph{adversarial example} of~$x_0$ if 
$C(x')~\neq~s$ and
$\dist{x'-x_0}_p \leq \varepsilon$ for a given distortion bound~$\varepsilon\in \RR^+$ and $l_p$ norm. 
The attacker searches for adversarial inputs~$x'$ with low distortion while maximizing a loss function $\mathcal{L}$, e.g., cross-entropy loss. Formally, the optimization problem is defined as follows:
\begin{equation}
\label{eq:opt_goal}
    \zeta = \argmax_{\lVert\zeta\rVert_p\leq\varepsilon}\mathcal{L}(y,x+\zeta,\theta)
\end{equation}

\subsection{Main Intuition}

Existing work on transferability~\cite{taig,chen2024rethinking,DBLP:conf/uss/DemontisMPJBONR19,transferstudysp} highlighted that adversarial examples transfer better when the surrogate model is similar to the target model. However, up until now, this similarity has mostly been investigated in terms of architecture types, ignoring the impact of other training parameters, such as the choice of optimizer, scheduler, and amount of available data.

We show in what follows that the explicit introduction of model heterogeneity by varying other parameters relevant for training, e.g., optimizer, scheduler, and amount of available data, is expected to reduce the overall transferability. We show that (1) the choice of model parameters, e.g., architecture, scheduler, optimizer, promotes gradient diversity and (2) transferability can be quantified with a gradient comparison of the surrogate and target model~\cite{DBLP:conf/uss/DemontisMPJBONR19}.

\begin{proposition}
    A model $f$ with major parameter configuration $\mathcal{P}$ and hyperparameter configuration $\mathcal{H}$ is optimized to model parameters $\theta$ during training. Heterogeneity, i.e., a change in these parameter configurations,  $\hat{\mathcal{P}}, \hat{\mathcal{H}}$, results in a model $\hat{f}$ (parameterized by $\hat{\theta}$) with a changed loss landscape and hence diverse (by a sufficiently large $\gamma$) model gradients for a set of input points $\mathcal{X}$. Concretely, 
    \begin{equation}
   \sum_{x\in\mathcal{X}}\lVert\nabla_x\mathcal{L}(y,x,\theta)-\nabla_x\mathcal{L}(y,x,\hat{\theta})\rVert_2 > \gamma
    \end{equation}
\end{proposition}
Recall that the supervised training procedure for a generic multi-class classifier $C$ aims at finding the optimal set of parameters $\theta_{\mathcal{P}, \mathcal{H}}$ given parameter configurations $\mathcal{P}, \mathcal{H}$ that minimize the aggregated loss over the entire training set $\mathcal{D}$: 
\begin{equation}
    \min_{\theta_{\mathcal{P}, \mathcal{H}}
}\sum_{(x,y)\in\mathcal{D}}^{}\mathcal{L}(y, x,     \theta_{\mathcal{P}, \mathcal{H}})
\end{equation}
with sample $x$, ground-truth $y$, and loss-function $\mathcal{L}$.

We expect that the loss landscape defined by the parameters $\theta$, i.e., the converged result of the non-convex optimization problem, depends on many factors, such as the amount of available data, i.e., the partition of data across \emph{N} nodes,  architecture, and potentially on the optimization strategy. Beyond that, the respective hyperparameters $\mathcal{H}$, i.e., learning rate \Lr, momentum \Momen, and weight decay \Wd, might also impact the convergence behavior of the optimization procedure.
As a result, a change in $\mathcal{P}$ and $\mathcal{H}$ converges to a different set of model parameters, i.e., $\hat{\theta}_{\hat{\mathcal{P}}, \hat{\mathcal{H}}}$. For clarity, we denote $\theta := \theta_{\mathcal{P}, \mathcal{H}}$ and $\hat{\theta} := \theta_{\hat{\mathcal{P}}, \hat{\mathcal{H}}}$. We empirically show an increase in gradient diversity with an increase in heterogeneity in $\mathcal{P}$ and $\mathcal{H}$ in the evaluation (cf. Table~\ref{tab:grad_sim}).

\begin{proposition}
    The transferability of adversarial examples from a surrogate model to a target model can be evaluated by comparing the model gradients for a given point using the cosine similarity function with:     
    \begin{equation}\label{eq:s}
    S(x,y) =\frac{\nabla_x{\mathcal{L}(y,x,\hat{\theta})}^\intercal\nabla_x{\mathcal{L}(y,x,{\theta})}}{\lVert\nabla_x{\mathcal{L}(y,x,\hat{\theta})}\rVert_2\lVert\nabla_x{\mathcal{L}(y,x,{\theta})}\rVert_2}
    \end{equation}
\end{proposition}
As the ultimate goal of a transfer-based attack is to evade a target model, we are interested in the loss $\mathcal{L}$ that an adversarial example crafted on a surrogate model (parameterized by $\hat{\theta}$), i.e., $x'=x+\hat{\zeta}$, can obtain on the target model (parameterized by $\theta$). This loss, namely ${\mathcal{L}(y,x+\hat{\zeta},\theta)}$, defines transferability. In practice, we can rewrite $\mathcal{L}$ with a linear approximation~\cite{DBLP:journals/corr/GoodfellowSS14} as: 
\begin{equation}
\label{eq:lineardecomp}
    \mathcal{L}(y,x+\hat{\zeta},\theta) \approx \mathcal{L}(y,x,\theta) + {\hat{\zeta}}^\intercal\nabla_x\mathcal{L}(y,x,\theta) 
\end{equation}
This is subsequently combined with the optimization goal in~\Cref{eq:opt_goal}. To maximize the second term, we maximize the inner product over an $\varepsilon$-sized sphere as follows, with $l_q$ being the dual norm of $l_p$:
\begin{equation}
\label{eq:maxit}
    \max_{\lVert\hat{\zeta}\rVert_p\leq\varepsilon}\hat{\zeta}^\intercal\nabla_x\mathcal{L}(y,x,\hat{\theta})=\varepsilon\lVert\nabla_x\mathcal{L}(y,x,\hat{\theta})\rVert_q
\end{equation}
To maximize~\Cref{eq:maxit} and subsequently also \Cref{eq:lineardecomp}, we insert an optimal value of $\hat{\zeta}=\varepsilon\frac{\nabla_x\mathcal{L}(y,x,\hat{\theta})}{\lVert\nabla_x\mathcal{L}(y,x,\hat{\theta})\rVert}_2$ for an $l_2$ norm. As a result, the change in loss under a transfer attack for such a point is defined as:
\begin{equation}
\label{eq:lossdiff}
    \Delta_\mathcal{L} = \varepsilon\frac{{\nabla_x{\mathcal{L}}(y,x,\hat{\theta})}^\intercal}{\lVert\nabla_x{\mathcal{L}}(y,x,\hat{\theta})\rVert_2}\nabla_x\mathcal{L}(y,x,{\theta})\leq\varepsilon\lVert\nabla_x\mathcal{L}(y,x,{\theta})\rVert_2
\end{equation}
The left-hand side of the equation reflects the black-box case, which is upper-bounded by the white-box case on the right-hand side. 
Rearranging~\Cref{eq:lossdiff} reveals that the change in loss and, hence the impact of a transfer-based attack crafted for a sample $x$ can be inferred by comparing the gradient similarity of the surrogate and target model evaluated at sample $x$ as follows:
\begin{equation}
\label{eq:gradsim}
    S(x,y) =\frac{\nabla_x{\mathcal{L}(y,x,\hat{\theta})}^\intercal\nabla_x{\mathcal{L}(y,x,{\theta})}}{\lVert\nabla_x{\mathcal{L}(y,x,\hat{\theta})}\rVert_2\lVert\nabla_x{\mathcal{L}(y,x,{\theta})}\rVert_2}
\end{equation}
This illustrates that the change in loss between the surrogate and target model for a given sample $x$, i.e., the transferability, depends on the gradient similarity of the respective models. As a result, a decrease in gradient similarity, i.e., an increase in model heterogeneity (cf. Proposition~1), decreases the obtainable change in loss on the target classifier with an adversarial example crafted on a surrogate classifier (cf. Proposition~2), i.e., reduces transferability.

\section{Experimental Approach}

\subsection{Training parameters}
Training a machine learning (ML) model involves numerous decisions, including selecting the model's architecture, such as VGG or DenseNet, and determining its width and depth, typically based on the complexity of the task at hand. Additionally, hyperparameters such as learning rate and momentum require careful tuning. This tuning often involves empirically testing various values within a defined range to identify the configuration that yields optimal convergence.

Additionally, while the standard stochastic gradient descent (SGD) optimizer is commonly used for smaller ML projects, various alternative optimizers, such as ADAM, Adagrad, and Rprop, have been developed to enhance convergence depending on the nature of the training data. 
Furthermore, research has demonstrated that starting with a high learning rate and subsequently decreasing it as the number of epochs progresses can lead to improved performance and accuracy~\cite{adam}. This approach has catalyzed the development of various learning rate schedulers, including StepLR and ExponentialLR, which automatically adjust the learning rate to maximize convergence.

In our experiments, we opted to separate the parameters into two groups: the major parameters, which encompass choices of architecture, optimizer, and scheduler, from the hyperparameters, which include the learning rate, momentum, and weight decay. 
As detailed in Table~\ref{tab:parameters}, we selected a set of eight different architectures (\Arch), nine different optimizers (\Optim), and five different schedulers (\Sched). These selections were derived from commonly used values in academic research and practical projects. For the choice of architectures, we simplified the selection process by only selecting one representative architecture for each family of architectures---such as VGG11 for the VGG family. 

Hyperparameters, on the other hand, are tailored to the specific combination of major parameters. For example, a learning rate (\Lr) of 0.1, although common with SGD, leads to divergence when combined with the Adam optimizer. Thus, we tuned the hyperparameters as follows.

\subsection{Parameter tuning}
To ensure the quality of the trained model, each weak learner undergoes a phase of hyperparameter tuning prior to the actual training process.
The exact process is summarized in \Cref{alg:weaklearner}.
Initially, each weak learner takes as input its local dataset $D$, and a set of common parameters \texttt{params}, and draws the remaining parameters (\texttt{diverseParams}) randomly from the set of possible values (Line~\ref{alg:weaklearner:forloop}). Each node then evaluates a wide range of hyperparameter values over two subsequent tuning rounds, leveraging the Ray Tune tool~\cite{raytune} to identify the optimal set of parameters (Line~\ref{alg:weaklearner:tune}).
Specifically. During the fine-tuning steps, we split the node’s training data into an 80\%-20\% ratio for the training and validation sets, respectively. Ray Tune is configured to run up to 100 experiments per tuning step to determine the most effective hyperparameters.
This approach ensures that each weak learner is finely tuned, enhancing the model's overall performance and robustness.
Once a node finishes tuning its local training parameters, it runs a complete training round for 200 epochs (Line~\ref{alg:weaklearner:train})
.

We argue that this training approach is more realistic than single-shot training based on a random selection of parameters. In most distributed ML settings, each node can expect to spend some effort ensuring that their local training achieves an acceptable level of accuracy.

\subsection{Impact of Parameter Diversity}
We devised four scenarios to answer our research questions and evaluate the impact of parameter diversity on transferability. 
Each scenario involves a varying number of nodes, denoted $\Nodes\in{3,5,7}$. Unless otherwise specified, each node is trained on its disjoint dataset drawn from the complete dataset following a uniform distribution.
\begin{itemize}
    \item \textbf{Ensemble Learning (ENS)} serves as our baseline, employing standard ensemble learning techniques. Here, we rely on publicly available code to train models to mitigate any potential bias.
    \item \textbf{Independent Tuning (IT)} shares the same major parameter among each weak learner but tunes its local hyperparameters independently based on its available data.
    \item \textbf{{\majorParam}-Parameter Diversity (D$_\majorParam$)} extends IT by introducing diversity to the parameter $\majorParam\in\{\Arch,\Optim,\Sched\}$. Similarly to IT, the remaining major parameters are the same for all the weak learners, and each weak learner performs a local tuning process based on their local dataset.
\end{itemize}
By systematically analyzing these scenarios, we aim to elucidate the effects of parameter diversity on the transferability of learned models across different contexts.
In all our evaluations, the probability vector of each weak learner is averaged, and the final classification is determined by selecting the class with the highest probability. We consider the impact of other voting methods, such as \emph{weighted voting} and \emph{hard voting}, in RQ4.

\newcommand{\FullLineComment}[1]{\Statex \texttt{/* #1 */}}
\begin{algorithm}
\footnotesize
\caption{Training phase for Weak Learners}
\label{alg:weaklearner}
\begin{algorithmic}[1]
\State \textbf{Input:} dataset $D$, common parameters \texttt{params}, major parameters that should be randomly selected $\texttt{diverseParams}$
\State \textbf{Output:} Trained weak learner model $\texttt{model}$\vspace{0.5em}
\FullLineComment{For each \majorParam that should be diversified, randomly select it from the list of valid parameters}
\For{p in diverseParams}
    \State\label{alg:weaklearner:forloop} params.p = random(p.possibleValues)
\EndFor \vspace{0.5em}
\FullLineComment{Local tuning step to derive the best hyperparameters \hyperParam: \Lr, \Momen, and \Wd}
\State\label{alg:weaklearner:tune} \hyperParam =\Lr, \Momen, \Wd = $\mathtt{Tune(train\_model, params, data)}$\vspace{0.5em}
\FullLineComment{Final training based on the selected \majorParam and tuned \hyperParam}
\State\label{alg:weaklearner:train} model = $\mathtt{train\_model(params, \hyperParam,data)}$
\State \textbf{return} model
\end{algorithmic}
\end{algorithm}

\begin{table}
    \centering
    \begin{adjustbox}{max width=\linewidth}
    \begin{tabular}{c|lcl}
        \toprule
        &\textbf{Parameter} & \textbf{Variable} &\textbf{Possible values} \\
        \midrule
        \multirow{14}{*}{\rotatebox[origin=c]{90}{\textbf{Major Parameters \majorParam}}} & \multirow{2}{*}{\shortstack[l]{Number of \\nodes}} &\multirow{2}{*}{\Nodes} & \multirow{2}{*}{1, 3, 5, 7}\\
        & & & \\
        \cmidrule{2-4}
        &\multirow{7}{*}{Architecture} & \multirow{7}{*}{\Arch} & \multirow{7}{*}{\shortstack[l]{VGG19, MobileNetv2,\\ EfficientNet\_b0, DenseNet121,\\ SimpleDLA, ResNet18,0\\ ResNext29\_2x64d, DPN92, \\SeNet18, googlenet, \\shufflenetg2, regnetx\_200mf, \\preactresnet18}} \\
        & & & \\
        & & & \\
        & & & \\
        & & & \\
        & & & \\
        & & & \\
        \cmidrule{2-4}
        &\multirow{3}{*}{Optimizer} &\multirow{3}{*}{\Optim } & SGD, SGD$_{momentum}$, Adam\\
        && & SGD$_{nesterov}$, NAdam, Adagrad, \\
        && & ASGD, Rprop, RMSprop \\
        \cmidrule{2-4}
        &\multirow{3}{*}{Scheduler} & \multirow{3}{*}{\Sched} & CosineAnnealingLR, StepLR, \\
        && & ExponentialLR, CyclicLR, \\
        && & ReduceLROnPlateau \\
        \midrule
        \multirow{3}{*}{\rotatebox[origin=c]{90}{\textbf{\shortstack{Hyper-\\params\\\hyperParam}}}} &Learning rate &\Lr & $[0.0001, 0.1]$ \\
        \cmidrule{2-4}
        &Momentum & \Momen& $[0, 0.99]$\\
        \cmidrule{2-4}
        \cmidrule{2-4}
        &Weight decay & \Wd & $[0.00001, 0.01]$ \\
        \bottomrule
    \end{tabular}
    \end{adjustbox}
    \caption{Parameters considered for the experiments.}
    \label{tab:parameters}
\end{table}

\section{Experiments}
\label{sec:experiments}
Let~$S\subset \setX \times \setY$ denote the set of (labeled) genuine samples provided to the attacker $\adv$, let~$n := \size{S}$,
and let~$\varepsilon$ denote the distortion budget.
To compute the attack success rate ($\mathsf{ASR}$) in practice, we determine the number of successful adversarial examples generated by the attacker:
\begin{equation}
n_{\mathrm{succ}} := \Biggl\vert
\left\{
  (x,x') \in \mathcal{X}\times\adv(S) \;\middle|\;
  \begin{aligned}
  & C(x)\neq C(x')\;\land \\
  & \dist{x'-x}_p \leq \varepsilon
  \end{aligned}
\right\}\Biggr\vert
\end{equation}
where~$\adv(S)$ denotes the set of candidate adversarial examples output by~$\adv$ in a run of the attack on input~$S$.
The $\mathsf{ASR}$ is defined as $\advsuccrate := \nicefrac{n_{\mathrm{succ}}}{n}$, i.e., the ratio of successful adversarial examples.
The complement of the $\advsuccrate$ is the robust accuracy ($\robustacc$) of the classifier, i.e., $\robustacc = 1 - \advsuccrate$.

\subsection{Experimental Setup}
All our experiments were run on an Ubuntu 24.04 machine featuring two NVIDIA A40 GPUs and one NVIDIA H100 GPU, two AMD EPYC 9554 64-core Processors, and 768 GB of RAM. All scenarios were executed using Python 3.9.18, CUDA 12.5, Pytorch 2.2.1, and Ray Tune 2.9.3.
Due to the limited amount of data available to each weak learner, we rely on well-known data augmentation techniques such as random flipping and random resize and cropping from the Torchvision transformsV2 library.

Due to space constraints, we include our full results and analysis for the CIFAR10 dataset in Table~\ref{tab:mainresults} and provide the main results for the FashionMNIST dataset in Table~\ref{tab:regression_complete}. We note, however, that most of our findings are consistent across both datasets.

All our results are averaged over five independent runs. Where appropriate, we also provide the 95\% confidence interval. 

\subsection{Selection of the Attacks}
We evaluate our setup against the state-of-the-art common weakness attack (CW)~\cite{chen2024rethinking}, which improves transferability by using multiple (an ensemble of) surrogate models, effectively outperforming existing attacks. We rely on the CW attack for the evaluation as we argue it is the best strategy for an adaptive adversary that knows that the inference model is distributed across many different architectures trained using disparate parameters. In addition to CW, we consider two derivative attacks: Sharpness Aware Minimization (SAM) and Cosine Similarity Encourager (CSE), which are defined in the same paper. For the sake of completeness, our evaluation includes all three variants.
Note that we do not evaluate attacks like  \cite{DBLP:conf/iclr/BryniarskiHPWC22}, which focus on multi-objective optimization challenges involved in fooling a target model while simultaneously circumventing defenses.
Since we evaluate the transferability on undefended models in our setup, we focus on the CW attack, owing to its superior transferability rate.

We define the maximum distortion bound $\varepsilon=8/255$. For each considered attack, we generate an adversarial example for each sample of the test dataset. For example, in CIFAR10, this results in \num{10000} adversarial examples per attack. 
\iflongversion
   For reproducibility purposes, we provide the complete list of parameters in Appendix~\ref{sec:reproducibility}.
\fi
\subsection{Evaluation Results}

We now proceed to answer our research questions (RQ1-RQ4) empirically using experiments in CIFAR10 and FashionMNIST.

\begin{figure}[tb]
    \centering
        \includegraphics[width=.45\textwidth]{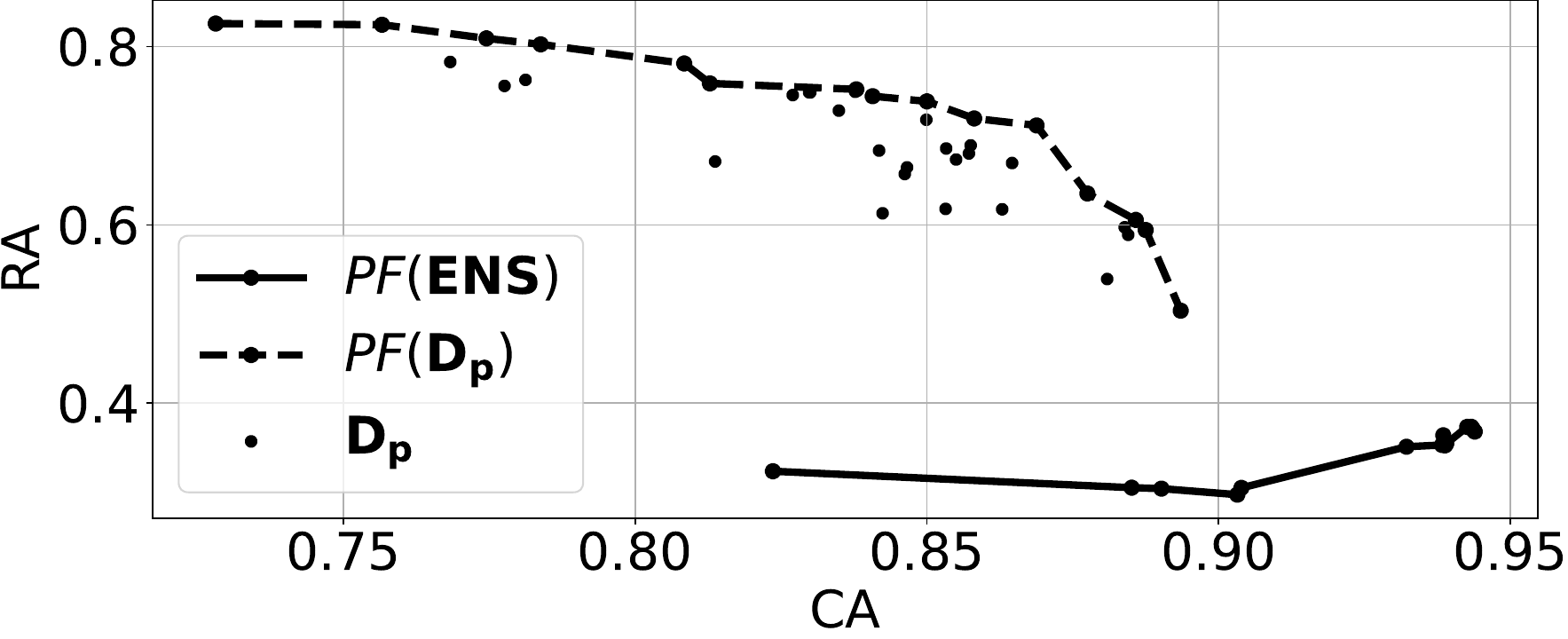}
    \caption{Pareto frontier of all accuracy-robustness tradeoffs of distributed instantiations (D$_\majorParam$) compared to the baseline ensemble (\textbf{ENS}).}
    \label{fig:pareto_frontier}
\end{figure}

\vspace{1 em}\noindent \textbf{RQ1: }
We leverage the Pareto frontier to illustrate the obtainable accuracy-robustness tradeoffs across all D$_\majorParam$ instantiations and our ensemble baseline (\textbf{ENS}). 
Recall that the Pareto frontier emerges as an effective tool to evaluate tradeoffs between the clean accuracy $\mainacc$ and the robust accuracy $\robustacc$.
A solution~$\omega^*$
is \emph{Pareto optimal} if there exists no other solution that improves all objectives simultaneously.
Formally, given two solutions~$\omega_1$ and~$\omega_2$, we write~$\omega_1 \succ \omega_2$ if~$\omega_1$ dominates~$\omega_2$, i.e., if
$\mainacc(\omega_1) \geq \mainacc(\omega_2) \land \robustacc(\omega_1) \geq \robustacc(\omega_2)$,  where $\mainacc(\omega)$ and~$\robustacc(\omega)$ denote the clean accuracy and robust accuracy as functions of the parameter~$\omega$. The \emph{Pareto frontier} is the set of Pareto-optimal solutions:
\begin{equation}
    PF(\Omega) = \{ \omega^*\in \Omega \ | \ \nexists \omega \in \Omega\text{ s.t. } \omega\succ \omega^*\}.
\end{equation}
This enables us to compare the average $\robustacc$ (and respective $\mainacc$) across all attacks, i.e., the last column of \Cref{tab:mainresults}, in~\Cref{fig:pareto_frontier}.

For the ensemble baseline, we observe a wide range of \mainacc between $82$ and $94$\%, with a narrow and hence limited range of \robustacc with a maximum of $37$\%. In stark contrast, the D$_\majorParam$ instances result in a broader range of tradeoffs. Concretely, we observe improvements in \robustacc of $20-40\%$, i.e., primarily between $60-80\%$ with a maximum \robustacc of up to $83$\%. While the prioritization of either \mainacc or \robustacc is use-case specific, we find a near-optimal point of operation for $\mainacc=87\%, \robustacc=71\%$.

\vspace{1 em}\noindent \textbf{RQ2: }To explore the impact of parameter diversity on robustness,
We rely on the standard Ordinary Least Squares regression method 
\iflongversion
   (cf. Appendix~\ref{sec:reg_analysis})
\fi
using the major parameters \majorParam, i.e., number of nodes (\Nodes), diversity in architecture (\Arch), optimizer (\Optim), scheduling (\Sched), and independent tuning (IT) as explanatory variables, to explain the robustness as the response variable. 
Our findings are detailed in Table~\ref{tab:regression_complete}, using standard ensemble models as the baseline. Our regression model demonstrates strong robustness, accounting for 87.4\% of the variability of the dependent variable through the selected predictors, indicated by an R-squared value of 0.874. Notably, the robustness improves by 2.88\% for each additional node. It is important to note that this linear model does not capture the eventual saturation effect, where robustness gains diminish as the size of \Nodes grows large.
Furthermore, our results show an increase of 34.53\% when the weak learners use diverse hyperparameters \hyperParam through independent tuning (IT), including the learning rate \Lr, momentum \Momen, and weight decay \Wd. As noticed in the ablation study, we note, however, that the diversity of the optimizer, architecture, and scheduler does not appear to have a statistically significant contribution to the robustness.

\vspace{1 em}\noindent \textbf{RQ3: }
To assess the impact of non-IID (non-independent and identically distributed) data distribution on both accuracy and robustness, we conduct independent tuning experiments under two distinct data partitioning schemes: uniform distribution and Dirichlet distribution. 
The Dirichlet distribution, commonly employed in many Federated Learning (FL) deployments~\cite{DBLP:conf/aistats/BagdasaryanVHES20,DBLP:conf/icdcs/AndreinaMMK21}, allows us to simulate more realistic, heterogeneous data environments. In the case of the Dirichlet distribution, the number of samples of a given class is distributed among the weak learners with parameter $\alpha=0.9$.

Our results (cf. Table~\ref{tab:data_dirichlet}) show that the accuracy of models trained with non-IID data decreases by 3.1\% for $\Nodes=3$ and 2.6\% for $\Nodes=7$ when compared to models trained with uniformly distributed data. On the other hand, the effect on the mean robustness is negligible, with an increase of only 0.2\% and 0.6\% when $\Nodes=3$ and $\Nodes=7$, hinting that non-IID data distribution among the nodes does not directly impact the robustness of the distributed model.

This observation is further substantiated by our regression analysis shown in Table~\ref{tab:regression_dirichlet}, using IT as the baseline and \Nodes and Dirichlet as the explanatory variables. Our analysis indicates that using the Dirichlet distribution does not produce statistically significant differences in robustness performance compared to the uniform distribution (i.e., the $P$-value for the $t$-test is high, supporting the null hypothesis).

\begin{table}

\begin{adjustbox}{max width=\linewidth}
\begin{tabular}{ll|c|lll|l}
\toprule
\multirow{2}{*}{Exp.} & \multirow{2}{*}{$N$} & \multirow{2}{*}{$\mainacc$} & \multicolumn{4}{c}{$\robustacc$}\\
\cmidrule{4-7}
&&& SAM & CSE& CW & Mean\\
\midrule
\multirow[t]{3}{*}{\textbf{IT}} & 3 & 0.87 {\tiny $\pm$ 0.01} & 0.60 {\tiny $\pm$ 0.05} & 0.61 {\tiny $\pm$ 0.04} & 0.59 {\tiny $\pm$ 0.04} & 0.60 \\
 \textbf{IID}& 5 & 0.84 {\tiny $\pm$ 0.01} & 0.71 {\tiny $\pm$ 0.04} & 0.70 {\tiny $\pm$ 0.02} & 0.69 {\tiny $\pm$ 0.03} & 0.70 \\
 & 7 & 0.81 {\tiny $\pm$ 0.01} & 0.77 {\tiny $\pm$ 0.04} & 0.74 {\tiny $\pm$ 0.03} & 0.73 {\tiny $\pm$ 0.03} & 0.74 \\
\midrule
\multirow[t]{3}{*}{\textbf{IT}} & 3 & 0.84 {\tiny $\pm$ 0.01} & 0.60 {\tiny $\pm$ 0.02} & 0.62 {\tiny $\pm$ 0.02} & 0.60 {\tiny $\pm$ 0.02} & 0.61 \\
 \textbf{Non-}& 5 & 0.83 {\tiny $\pm$ 0.01} & 0.70 {\tiny $\pm$ 0.03} & 0.70 {\tiny $\pm$ 0.03} & 0.68 {\tiny $\pm$ 0.03} & 0.69 \\
\textbf{IID} & 7 & 0.79 {\tiny $\pm$ 0.02} & 0.77 {\tiny $\pm$ 0.03} & 0.74 {\tiny $\pm$ 0.02} & 0.73 {\tiny $\pm$ 0.02} & 0.75 \\

\bottomrule
\end{tabular}
\end{adjustbox}
\caption{Impact of data distribution on the accuracy and robustness in the independent training scenario. }
\label{tab:data_dirichlet}
\end{table}

\begin{table}
\begin{adjustbox}{max width=\linewidth}
\begin{tabular}{lcccccc}
\toprule
                   & \textbf{coef} & \textbf{std err} & \textbf{t} & \textbf{P $> |$t$|$} & \textbf{[0.025} & \textbf{0.975]}  \\
\midrule
\textbf{const}     &       0.5097  &        0.039     &    13.237  &         {0.000}        &        0.431    &        0.589     \\
\textbf{N}         &       0.0354  &        0.007     &     5.064  &         {0.000}        &        0.021    &        0.050     \\
\textbf{\textbf{Dirichlet}} &       0.0042  &        0.023     &     0.184  &         0.855        &       -0.043    &        0.051     \\
\bottomrule
\end{tabular}

\end{adjustbox}
\caption{Regression analysis of the impact of the data distribution on the robustness of the models in the independent training scenario.}
\label{tab:regression_dirichlet}
\end{table}

\vspace{1 em}\noindent \textbf{RQ4: }
Last, we examine the impact of different voting schemes on the accuracy and robustness of distributed ML. We consider the following three schemes:
\begin{enumerate}
    \item \textbf{Average voting}: In this scheme, the output vectors of each model are averaged, and the class with the highest average probability is selected as the final prediction. This has been the baseline scheme used in our previous experiments.
    \item \textbf{Hard voting}: Each weak learner votes for the class with the highest probability according to its local model. The class that receives the majority of votes across all learners is chosen as the final prediction.
    \item \textbf{Weighted voting}: Similar to hard voting, each weak learner votes for the class with the highest probability, but the votes are weighted by the confidence level of the learner's model.
\end{enumerate}
As shown in Table~\ref{tab:data_voting}, the average voting scheme performs best compared to the other two schemes. Specifically, hard voting decreases accuracy by 3\% for $\Nodes=3$ and 2\% for $\Nodes=7$, with a corresponding decrease in mean robustness of 2\% and 1\%. Weighted voting results in smaller decreases, with accuracy decreasing by 1\% for  $\Nodes=3$ and  $\Nodes=7$, and robustness by 1\% for  $\Nodes=3$ and  $\Nodes=7$. This is further confirmed by our regression analysis in the last two rows of  Table~\ref{tab:regression_complete}; on average, the hard and weighted voting schemes result in a performance decrease of 1.1\% and 0.4\%. However, these impacts are not statistically significant, with P-values of 0.08 and 0.49, both above the 0.05 threshold.

\begin{table}[!ht]
\begin{adjustbox}{max width=\linewidth}
\begin{tabular}{ll|c|lll|l}
\toprule
\multirow{2}{*}{Voting} & \multirow{2}{*}{$N$} & \multirow{2}{*}{$\mainacc$} & \multicolumn{4}{c}{$\robustacc$}\\
\cmidrule{4-7}
&&& SAM & CSE& CW & Mean\\
\midrule
\multirow[t]{3}{*}{Average} & 3 & 0.86{ \tiny $\pm$ 0.03} & 0.64{ \tiny $\pm$ 0.06} & 0.65{ \tiny $\pm$ 0.05} & 0.63{ \tiny $\pm$ 0.04} & 0.64{ \tiny $\pm$ 0.05} \\
 & 5 & 0.83{ \tiny $\pm$ 0.03} & 0.72{ \tiny $\pm$ 0.06} & 0.71{ \tiny $\pm$ 0.04} & 0.69{ \tiny $\pm$ 0.05} & 0.70{ \tiny $\pm$ 0.05} \\
 & 7 & 0.80{ \tiny $\pm$ 0.02} & 0.79{ \tiny $\pm$ 0.03} & 0.76{ \tiny $\pm$ 0.03} & 0.75{ \tiny $\pm$ 0.03} & 0.76{ \tiny $\pm$ 0.03} \\
\midrule
\multirow[t]{3}{*}{Hard} & 3 & 0.83{ \tiny $\pm$ 0.05} & 0.63{ \tiny $\pm$ 0.06} & 0.63{ \tiny $\pm$ 0.05} & 0.61{ \tiny $\pm$ 0.04} & 0.62{ \tiny $\pm$ 0.05} \\
 & 5 & 0.82{ \tiny $\pm$ 0.03} & 0.71{ \tiny $\pm$ 0.06} & 0.69{ \tiny $\pm$ 0.04} & 0.68{ \tiny $\pm$ 0.05} & 0.69{ \tiny $\pm$ 0.05} \\
 & 7 & 0.78{ \tiny $\pm$ 0.03} & 0.78{ \tiny $\pm$ 0.03} & 0.74{ \tiny $\pm$ 0.02} & 0.73{ \tiny $\pm$ 0.03} & 0.75{ \tiny $\pm$ 0.03} \\
\midrule
\multirow[t]{3}{*}{Weighted} & 3 & 0.85{ \tiny $\pm$ 0.03} & 0.64{ \tiny $\pm$ 0.06} & 0.64{ \tiny $\pm$ 0.05} & 0.62{ \tiny $\pm$ 0.04} & 0.63{ \tiny $\pm$ 0.05} \\
 & 5 & 0.83{ \tiny $\pm$ 0.03} & 0.71{ \tiny $\pm$ 0.06} & 0.70{ \tiny $\pm$ 0.05} & 0.69{ \tiny $\pm$ 0.05} & 0.70{ \tiny $\pm$ 0.05} \\
 & 7 & 0.79{ \tiny $\pm$ 0.02} & 0.79{ \tiny $\pm$ 0.04} & 0.75{ \tiny $\pm$ 0.03} & 0.74{ \tiny $\pm$ 0.03} & 0.76{ \tiny $\pm$ 0.03} \\\bottomrule
\end{tabular}
\end{adjustbox}
\caption{Accuracy and robustness of \textbf{D$_\Optim$} across the different voting schemes.}
\label{tab:data_voting}
\end{table}

\begin{table}[tbp]
\begin{adjustbox}{max width=\linewidth}
\begin{tabular}{c|lcccccc}
\toprule
       &             & \textbf{coef} & \textbf{std err} & \textbf{t} & \textbf{P $> |$t$|$} & \textbf{[0.025} & \textbf{0.975]}  \\
\midrule
\multirow{8}{*}{\rotatebox[origin=c]{90}{CIFAR10}}  &\textbf{const}        &       0.2058  &        0.012     &    16.924  &         {0.000}        &        0.182    &        0.230     \\
&\textbf{N}            &       0.0288  &        0.002     &    17.023  &         {0.000}        &        0.025    &        0.032     \\
&\textbf{\textbf{IT}} &       0.3453  &        0.010     &    35.539  &         {0.000}        &        0.326    &        0.364     \\
&\textbf{\textbf{D$_\Optim$}} &      -0.0012  &        0.006     &    -0.208  &         0.836        &       -0.013    &        0.010     \\
&\textbf{\textbf{D$_\Arch$}}  &       0.0076  &        0.006     &     1.271  &         0.204        &       -0.004    &        0.019     \\
&\textbf{\textbf{D$_\Sched$}} &      -0.0089  &        0.006     &    -1.498  &         0.135        &       -0.020    &        0.003     \\
&\textbf{Hard}         &      -0.0117  &        0.007     &    -1.723  &         0.086        &       -0.025    &        0.002     \\
&\textbf{Weighted}     &      -0.0046  &        0.007     &    -0.679  &         0.498        &       -0.018    &        0.009     \\
\midrule 
\multirow{8}{*}{\rotatebox[origin=c]{90}{FashionMNIST}}  &\textbf{const}        &       0.9403  &        0.003     &   306.244  &         {0.000}        &        0.934    &        0.946     \\
&\textbf{N}            &       0.0030  &        0.000     &     6.692  &         {0.000}        &        0.002    &        0.004     \\
&\textbf{\textbf{IT}} &       0.0026  &        0.004     &     0.723  &         0.471        &       -0.005    &        0.010     \\
&\textbf{\textbf{D$_\Optim$}} &       0.0005  &        0.002     &     0.269  &         0.788        &       -0.003    &        0.004     \\
&\textbf{\textbf{D$_\Arch$}}  &       0.0015  &        0.002     &     0.642  &         0.522        &       -0.003    &        0.006     \\
&\textbf{\textbf{D$_\Sched$}} &      -0.0014  &        0.002     &    -0.577  &         0.565        &       -0.006    &        0.003     \\
&\textbf{Hard}         &      -0.0063  &        0.002     &    -3.489  &         {0.001}        &       -0.010    &       -0.003     \\
&\textbf{Weighted}     &      -0.0048  &        0.002     &    -2.652  &         {0.009}        &       -0.008    &       -0.001     \\
\bottomrule
\end{tabular}
\end{adjustbox}
\caption{Complete regression analysis of the impact of the different parameters and voting scheme on the average robustness of the distributed models.}
\label{tab:regression_complete}
\end{table}

\begin{table}
\centering
\begin{adjustbox}{max width=\linewidth}
\begin{tabular}{llll}
\toprule
                                                & \textbf{$N=3$}              & \textbf{$N=5$} & \textbf{$N=7$}   \\
\midrule
S    & 0.028  &       0.029  &        0.020        \\
\robustacc &   0.66 &       0.67  &        0.73        \\
\bottomrule
\end{tabular}

\end{adjustbox}
\caption{Gradient similarity and \robustacc in relation with the ensemble \textbf{ENS} and D$_{\Arch\Optim\Sched}$.}
\label{tab:grad_sim}
\end{table}

\subsection{Ablation Study} 

Our results are summarized in Table~\ref{tab:mainresults}, from which several key insights can be derived. 
Notably, we observe a consistent decline in accuracy as the number of nodes (\Nodes) increases in the distributed ML scenarios (denoted as the D$_{\Arch\Optim\Sched}$ rows in the table). Here, the accuracy slightly declines from 85\% down to 82\%. 
This degradation in performance can be primarily attributed to the division of the dataset among the weak learners, which reduces the amount of data available for each learner as the node count increases.
In contrast, standard ensemble models (represented by the \textbf{ENS} rows) do not exhibit this limitation, as each model in the ensemble has access to the complete dataset.

Our results indicate a continuous improvement as \Nodes increases across ensemble and distributed ML models. For the baseline ensemble scenario, the mean robustness increases from 31\% to 37\% as \Nodes increases from 3 to 7. This trend is similar in the distributed ML system, where the robustness increases from 66\% to 73\%, an increase of 7\% compared to standard ensemble models.

Last but not least, we selectively measure the gradient similarity (cf. \Cref{eq:s}) on a subset of our results with the surrogate models in \Cref{tab:grad_sim}. For instance, for the case of $D_{\Arch\Optim\Sched}$ in CIFAR10, our results show that (1) gradient similarity (compared to \textbf{ENS}) decreases for diverse models as \Nodes increases from 5 to 7 (cf. Proposition 1) and (2) gradient similarity follows a similar trend to transferability, i.e., \advsuccrate, (which is inversely related to \robustacc) for diverse models as \Nodes increases (cf. Proposition 2). 
\iflongversion
   We provide the complete gradient similarity analysis of all the scenarios in Appendix~\ref{sec:heatmap}.
\else
The complete gradient analysis can be found in the full version of our paper~\cite{arxiv}.
\fi

\begin{table}[tbp]

\begin{adjustbox}{max width=\linewidth}
\begin{tabular}{ll|c|lll|l}
\toprule
\multirow{2}{*}{Exp.} & \multirow{2}{*}{$N$} & \multirow{2}{*}{$\mainacc$} & \multicolumn{4}{c}{$\robustacc$}\\
\cmidrule{4-7}
&&& SAM & CSE& CW & Mean\\
\midrule
\multirow[t]{3}{*}{\textbf{ENS}} & 3 & 0.88 {\tiny $\pm$ 0.03} & 0.26 {\tiny $\pm$ 0.02} & 0.34 {\tiny $\pm$ 0.01} & 0.31 {\tiny $\pm$ 0.0} & 0.31 \\
 & 5 & 0.94 {\tiny $\pm$ 0.0} & 0.29 {\tiny $\pm$ 0.01} & 0.4 {\tiny $\pm$ 0.01} & 0.37 {\tiny $\pm$ 0.0} & 0.35 \\
 & 7 & 0.94 {\tiny $\pm$ 0.0} & 0.29 {\tiny $\pm$ 0.0} & 0.43 {\tiny $\pm$ 0.0} & 0.39 {\tiny $\pm$ 0.0} & 0.37 \\
\midrule
\multirow[t]{3}{*}{\textbf{D$_{\Arch\Optim\Sched}$}} & 3 & 0.85 {\tiny $\pm$ 0.02} & 0.66 {\tiny $\pm$ 0.05} & 0.67 {\tiny $\pm$ 0.03} & 0.65 {\tiny $\pm$ 0.03} & 0.66 \\
 & 5 & 0.85 {\tiny $\pm$ 0.02} & 0.68 {\tiny $\pm$ 0.04} & 0.68 {\tiny $\pm$ 0.03} & 0.66 {\tiny $\pm$ 0.03} & 0.67  \\
 & 7 & 0.82 {\tiny $\pm$ 0.01} & 0.75 {\tiny $\pm$ 0.03} & 0.73 {\tiny $\pm$ 0.02} & 0.72 {\tiny $\pm$ 0.02} & 0.73  \\
\midrule
\midrule
\multirow[t]{3}{*}{\textbf{D$_\Arch$}} & 3 & 0.85 {\tiny $\pm$ 0.03} & 0.66 {\tiny $\pm$ 0.06} & 0.66 {\tiny $\pm$ 0.04} & 0.65 {\tiny $\pm$ 0.04} & 0.66 \\
 & 5 & 0.84 {\tiny $\pm$ 0.02} & 0.73 {\tiny $\pm$ 0.06} & 0.72 {\tiny $\pm$ 0.04} & 0.71 {\tiny $\pm$ 0.05} & 0.72 \\
 & 7 & 0.82 {\tiny $\pm$ 0.03} & 0.78 {\tiny $\pm$ 0.03} & 0.75 {\tiny $\pm$ 0.01} & 0.74 {\tiny $\pm$ 0.01} & 0.76 \\
\midrule
\multirow[t]{3}{*}{\textbf{D$_\Optim$}} & 3 & 0.86 {\tiny $\pm$ 0.03} & 0.64 {\tiny $\pm$ 0.06} & 0.65 {\tiny $\pm$ 0.05} & 0.63 {\tiny $\pm$ 0.04} & 0.64 \\
 & 5 & 0.83 {\tiny $\pm$ 0.03} & 0.72 {\tiny $\pm$ 0.06} & 0.71 {\tiny $\pm$ 0.04} & 0.69 {\tiny $\pm$ 0.05} & 0.70 \\
 & 7 & 0.80 {\tiny $\pm$ 0.02} & 0.79 {\tiny $\pm$ 0.03} & 0.76 {\tiny $\pm$ 0.03} & 0.75 {\tiny $\pm$ 0.03} & 0.76 \\
\midrule
\multirow[t]{3}{*}{\textbf{D$_\Sched$}} & 3 & 0.87 {\tiny $\pm$ 0.04} & 0.59 {\tiny $\pm$ 0.1} & 0.61 {\tiny $\pm$ 0.08} & 0.60 {\tiny $\pm$ 0.08} & 0.60 \\
 & 5 & 0.84 {\tiny $\pm$ 0.03} & 0.72 {\tiny $\pm$ 0.07} & 0.72 {\tiny $\pm$ 0.06} & 0.70 {\tiny $\pm$ 0.06} & 0.71 \\
 & 7 & 0.81 {\tiny $\pm$ 0.04} & 0.76 {\tiny $\pm$ 0.09} & 0.73 {\tiny $\pm$ 0.07} & 0.72 {\tiny $\pm$ 0.07} & 0.74 \\
\midrule
\multirow[t]{3}{*}{\textbf{D$_{\Arch\Optim}$}} & 3 & 0.87 {\tiny $\pm$ 0.01} & 0.61 {\tiny $\pm$ 0.02} & 0.63 {\tiny $\pm$ 0.02} & 0.61 {\tiny $\pm$ 0.02} & 0.62 \\
 & 5 & 0.84 {\tiny $\pm$ 0.03} & 0.70 {\tiny $\pm$ 0.04} & 0.70 {\tiny $\pm$ 0.02} & 0.68 {\tiny $\pm$ 0.03} & 0.69 \\
 & 7 & 0.79 {\tiny $\pm$ 0.04} & 0.80 {\tiny $\pm$ 0.05} & 0.76 {\tiny $\pm$ 0.03} & 0.75 {\tiny $\pm$ 0.04} & 0.77 \\
\midrule
\multirow[t]{3}{*}{\textbf{D$_{\Arch\Sched}$}} & 3 & 0.85 {\tiny $\pm$ 0.05} & 0.65 {\tiny $\pm$ 0.1} & 0.66 {\tiny $\pm$ 0.06} & 0.64 {\tiny $\pm$ 0.07} & 0.65 \\
 & 5 & 0.85 {\tiny $\pm$ 0.02} & 0.72 {\tiny $\pm$ 0.05} & 0.71 {\tiny $\pm$ 0.03} & 0.70 {\tiny $\pm$ 0.03} & 0.71 \\
 & 7 & 0.82 {\tiny $\pm$ 0.04} & 0.77 {\tiny $\pm$ 0.07} & 0.73 {\tiny $\pm$ 0.06} & 0.73 {\tiny $\pm$ 0.06} & 0.74 \\
\midrule
\multirow[t]{3}{*}{\textbf{D$_{\Optim\Sched}$}} & 3 & 0.86 {\tiny $\pm$ 0.02} & 0.62 {\tiny $\pm$ 0.1} & 0.63 {\tiny $\pm$ 0.07} & 0.61 {\tiny $\pm$ 0.08} & 0.62 \\
 & 5 & 0.85 {\tiny $\pm$ 0.01} & 0.71 {\tiny $\pm$ 0.06} & 0.70 {\tiny $\pm$ 0.04} & 0.69 {\tiny $\pm$ 0.04} & 0.70 \\
 & 7 & 0.80 {\tiny $\pm$ 0.02} & 0.77 {\tiny $\pm$ 0.05} & 0.75 {\tiny $\pm$ 0.04} & 0.73 {\tiny $\pm$ 0.04} & 0.75 \\
\bottomrule
\end{tabular}
\end{adjustbox}
\caption{Accuracy and robustness of the models trained in the different scenarios.}
\label{tab:mainresults}
\end{table}

\section{Conclusion}

In this work, we show that properly
distributed ML instantiations achieve across-the-board
improvements in accuracy-robustness tradeoffs against
state-of-the-art transfer-based attacks that could other-
wise not be realized by the current ensemble or federated
learning instantiations. Our results suggest that 
increasing the number of nodes and diversifying hyperparameters, such as the learning rate, the 
momentum and weight decay in distributed ML deployments play an important role in increasing overall robustness against transfer attacks. Surprisingly, our results suggest that other aspects, such as diverse architectures, optimizers, and schedulers, have little impact on robustness. We therefore hope that our findings motivate further research in this fascinating area.

\section{Acknowledgments}
This work has been co-funded by the Deutsche Forschungsgemeinschaft (DFG, German Research Foundation) under Germany’s Excellence Strategy - EXC 2092 CASA - 390781972, by the German Federal Ministry of Education and Research (BMBF) through the project TRAIN (01IS23027A), and by the European Commission through the HORIZON-JU-SNS-2022 ACROSS project (101097122). Views and opinions expressed are, however, those of the authors only and do not necessarily reflect those of the European Union. Neither the European Union nor the granting authority can be held responsible for them.

\bibliography{references.bib}

\appendix

\section{Regression Analysis}
\label{sec:reg_analysis}
In our paper, we relied on the Ordinary Least Squares regression method to extract the correlation between our explanatory variables, namely the diversity of the parameters, and the response variable, the robustness of the trained models. 
Our regression is obtained by minimizing the sum of the squared differences between the observed and predicted values by the linear model, ensuring the best-fitting line is found. Our regression model is specified as follows:
\begin{equation}
Y = \beta_0 + \beta_1X_1 + \beta_2X_2+...+\beta_kX_k+\epsilon,
\end{equation}
where $Y$ is the dependent variable, $X_1,...,X_k$ are the explanatory variables, $\beta_0$ is the intercept (this represents the baseline in our regressions), $\beta_1,...,\beta_k$ are the coefficients of the explanatory variables, and $\epsilon$ is the error term.
The coefficients are computed such that the residual sum of squares (RSS) is minimized, where $RSS=\sum_{i=1}^n(Y_i-\hat{Y_i})^2$ and $\hat{Y_i}$ represents the predicted value of $Y_i$.

Regressions help us understand the magnitude and direction of each explanatory variable's impact on the response variable. We further assess the statistical significance of our estimation by looking at the $t$-values and $P$-values associated with each coefficient. The $t$-value determines whether a coefficient in the regression model is significantly different from zero. It is computed as:
$t_i=\frac{\beta_i}{\sqrt{\hat\sigma^2(X'X)^{-1}_{ii}}}$, where the variance $\hat\sigma^2= \frac{\sum_{i=1}^n (y_i-\hat y_i)^2}{n-k-1}$, with $n$ the number of observations, and $k$  the number of explanatory variables considered in the regression.
On the other hand, the $P$-value measures the probability that the $t$-value would occur if the null hypothesis were true. It represents the probability that, due to randomness, we measure a certain $t$-value for a coefficient $\beta>0$ when the actual coefficient is 0. The $P$-value is derived from the $t$-value, the degrees of freedom (depending on the sample size), and the number of explanatory variables considered, using the Student's t-distribution. 
A low $P$-value (typically $\leq 0.05$) suggests strong evidence against the null hypothesis, meaning the coefficient is statistically significant and different from zero.
Conversely, a high $P$-value signifies the presence of little evidence against the null hypothesis, and the variable is not statistically significant from the baseline.

\section{Gradient Similarity Analysis}
\label{sec:heatmap}

\begin{figure}[htb!]
    \centering
    \includegraphics[width=.45\textwidth]{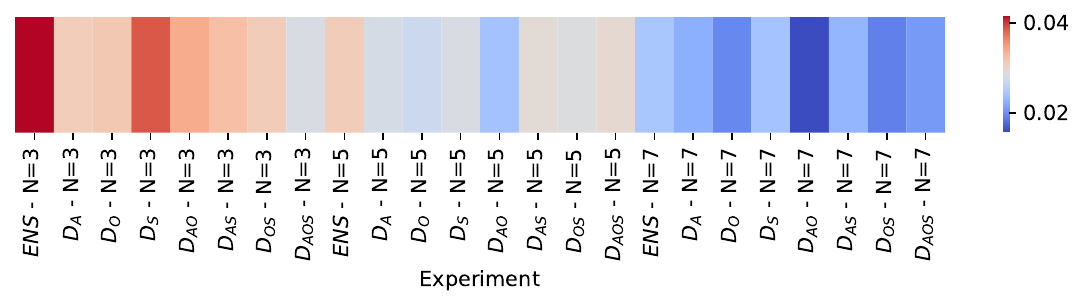}
    \caption{Heatmap of the similarity of individual models within our different scenarios with the surrogate models.}
    \label{fig:heatmap}
\end{figure}

We compare the similarity of individual models in our different setups with the surrogate models considered. To this end, we compute the gradient of each test dataset image in the surrogate and the weak learners' models and subsequently compare these gradients using the cosine similarity metric:
\begin{equation*}
    S(x,y) =\frac{\nabla_x{\mathcal{L}(y,x,\hat{\theta})}^\intercal\nabla_x{\mathcal{L}(y,x,{\theta})}}{\lVert\nabla_x{\mathcal{L}(y,x,\hat{\theta})}\rVert_2\lVert\nabla_x{\mathcal{L}(y,x,{\theta})}\rVert_2}
\end{equation*}

Recall that the cosine similarity, $S(x,y)$, quantifies the alignment between the gradient vectors, where $\mathcal{L}(y,x,\hat{\theta})$ and $\mathcal{L}(y,x,{\theta})$ represent the loss function gradients of the surrogate and weak learners' models, respectively. We then average these similarity scores across all test images to obtain an overall similarity measure for each model configuration in relation to the surrogate models.

Figure~\ref{fig:heatmap} illustrates the gradient similarity between the surrogate models and the target models. In line with our findings from Section~\ref{sec:experiments}, there is no distinct trend regarding the diversity of different model parameters. However, and consistent with our results, there is a noticeable decline in cosine similarity that can be observed as the number of nodes, denoted by $N$, increases.

\section{Parameters for Reproducibility}
\label{sec:reproducibility}
\begin{table*}[ht!]
    \centering
    \begin{adjustbox}{max width=\linewidth}
    \begin{tabular}{c|l|l|l|l|l|l}
        \toprule
        &\textbf{Parameter} &\textbf{Master \#1}  &\textbf{Master \#2}  &\textbf{Master \#3}  &\textbf{Master \#4}  &\textbf{Master \#5} \\
        \midrule
        \multirow{4}{*}{\rotatebox[origin=c]{90}{\textbf{\shortstack{Major \\Params \majorParam}}}} & {\shortstack[l]{Number of nodes \Nodes}} & 1 & 1 & 1 & 1 & 1 \\
        \cmidrule{2-7}
        &{Architecture \Arch} & VGG19 &  regnetx\_200mf & SimpleDLA & googlenet& DenseNet121 \\
        \cmidrule{2-7}
        &{Optimizer \Optim} & SGD & SGD$_{momentum}$& SGD$_{nesterov}$ &NAdam& Adam\\
        \cmidrule{2-7}
        &{Scheduler \Sched} &  CyclicLR & ReduceLROnPlateau & CosineAnnealingLR & ExponentialLR & StepLR\\
        \midrule
        \multirow{3}{*}{\rotatebox[origin=c]{90}{\textbf{\shortstack{Hyper-\\params\\\hyperParam}}}} &Learning Rate \Lr & 0.008256775377937505 & 0.006047937854649695 & 0.03264578338955776 & 0.0005200352529886811 & 0.000692270602808533\\
        \cmidrule{2-7}
        &Momentum \Momen & 0.7206164940541036 & 0.9858684617591867 & 0.8722845131660353 & 0.0 &  0.0\\
        \cmidrule{2-7}
        &Weight decay \Wd & 2.6053233480893608e-05 & 0.00010394001588929032 & 1.9589484425998513e-05 & 0.0004129327748482861 & 1.5842161316526423e-05   \\
\midrule
        & Test accuracy & 0.9005 & 0.9262 &        0.9227 & 0.9101 & 0.9094 \\
        \bottomrule
    \end{tabular}
    \end{adjustbox}
    \caption{Value considered for the underlying model parametrizations used in the experiments.}
    \label{tab:parameters_masterconfigs}
\end{table*}

In our experiments, we generated five baseline models derived by randomly sampling the major parameters from the list of possible values. These major parameters were iteratively redrawn until the test accuracy exceeded 90\%, ensuring a robust initial performance. Hyperparameters were subsequently fine-tuned using Ray Tune. A comprehensive list of all parameters that we utilized is detailed in Table~\ref{tab:parameters_masterconfigs}.

In our experimental scenarios, we always start with a master configuration and modify specific parameters accordingly. For instance, in the D$_\Optim$ scenario, we start with the configuration of Master~\#1, which specifies the use of VGG19 as the architecture and CyclicLR as the scheduler. We then introduce variability by randomly altering the optimizer \Optim parameter, thereby creating a diverse optimizer scenario. All scenarios are repeated five times, once for each master configuration.

\end{document}